\documentclass[sigconf]{acmart}
\usepackage[ruled,vlined,linesnumbered]{algorithm2e}
\usepackage{subcaption}
\usepackage{enumitem}
\usepackage{bbm}

\AtBeginDocument{%
  \providecommand\BibTeX{{%
    \normalfont B\kern-0.5em{\scshape i\kern-0.25em b}\kern-0.8em\TeX}}}

\copyrightyear{2024}
\acmYear{2024}
\setcopyright{acmlicensed}\acmConference[KDD '24]{Proceedings of the 30th ACM SIGKDD Conference on Knowledge Discovery and Data Mining}{August 25--29, 2024}{Barcelona, Spain}
\acmBooktitle{Proceedings of the 30th ACM SIGKDD Conference on Knowledge Discovery and Data Mining (KDD '24), August 25--29, 2024, Barcelona, Spain}
\acmDOI{10.1145/3637528.3671737}
\acmISBN{979-8-4007-0490-1/24/08}

\usepackage{amsthm}
\theoremstyle{plain}
\newtheorem{thm}{Theorem}[section] 

\theoremstyle{definition}

\begin{document}

\title{ Self-consistent Deep Geometric Learning for Heterogeneous Multi-source Spatial Point Data Prediction }

\author{Dazhou Yu}
\email{dyu62@emory.edu}
\orcid{0000-0003-2082-0834}
\affiliation{%
  \institution{Emory University}
  \city{Atlanta}
  \state{GA}
  \country{United States}
}

\author{Xiaoyun Gong}
\orcid{0000-0002-2761-5684}
\email{kristinagxy51@gmail.com}
\affiliation{%
  \institution{Emory University}
  \city{Atlanta}
  \state{GA}
  \country{United States}
}

\author{Yun Li}
\email{yli230@emory.edu}
\orcid{0000-0002-3205-8464}
\affiliation{%
  \institution{Emory University}
  \city{Atlanta}
  \state{GA}
  \country{United States}
}

\author{Meikang Qiu}
\email{qiumeikang@yahoo.com}
\orcid{0000-0002-1004-0140}
\affiliation{%
  \institution{Augusta University}
  \city{Augusta}
  \state{GA}
  \country{United States}
}

\author{Liang Zhao}
\email{liang.zhao@emory.edu}
\orcid{0000-0002-2648-9989}
\affiliation{%
  \institution{Emory University}
  \city{Atlanta}
  \state{GA}
  \country{United States}
}


\begin{abstract}
Multi-source spatial point data prediction is crucial in fields like environmental monitoring and natural resource management, where integrating data from various sensors is the key to achieving a holistic environmental understanding.
Existing models in this area often fall short due to their domain-specific nature and lack a strategy for integrating information from various sources in the absence of ground truth labels. Key challenges include evaluating the quality of different data sources and modeling spatial relationships among them effectively.
Addressing these issues, we introduce an innovative multi-source spatial point data prediction framework that adeptly aligns information from varied sources without relying on ground truth labels. A unique aspect of our method is the 'fidelity score,' a quantitative measure for evaluating the reliability of each data source. Furthermore, we develop a geo-location-aware graph neural network tailored to accurately depict spatial relationships between data points.
Our framework has been rigorously tested on two real-world datasets and one synthetic dataset. The results consistently demonstrate its superior performance over existing state-of-the-art methods.
\end{abstract}





\maketitle
\section{Introduction}
\begin{figure}
    \centering
    \includegraphics[width=\linewidth]{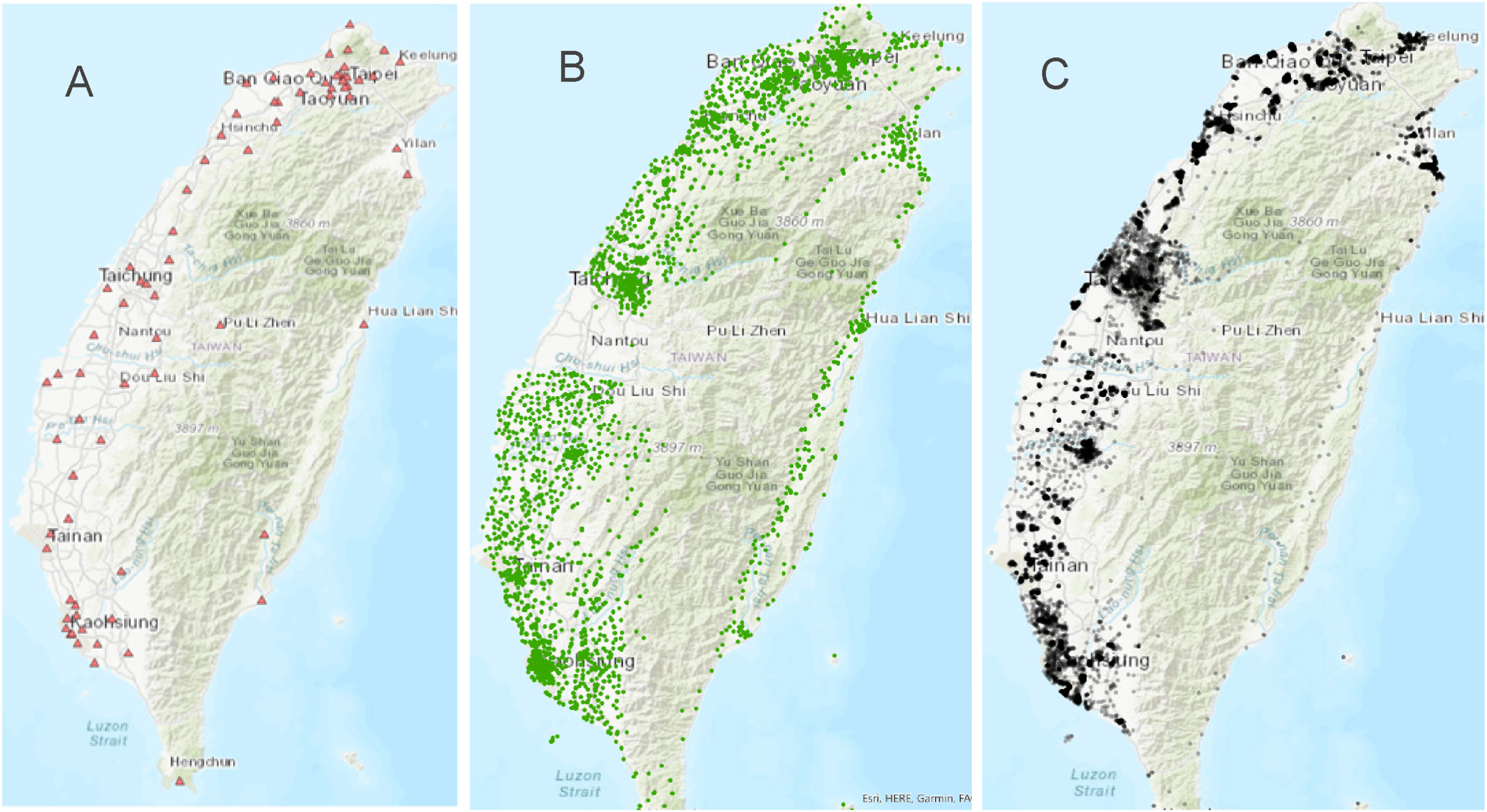}
    \caption{An example of multi-source spatial point prediction problem: Varying distribution of three data sources including (A) 74 AQMSs, (B) 3704 LASS AirBox sensors, and (C) 9701 EPA MicroStations.}
    \label{fig:intro}
\end{figure}

Spatial prediction \cite{zhang2023deep,yu2024stes,zhang2022deep} is essential in various domains such as environmental monitoring, natural resource management, and transportation planning \cite{jiang2018survey, hackett1990multi}. The process involves collecting various attributes and utilizing them to fit predictive models for target variables.  
PM 2.5 prediction is a popular application of spatial point prediction, where the data used for model training and prediction could be emission data or observations from fixed sites of air quality monitoring stations (AQMS).
Although AQMSs offer high-quality data, their deployment and maintenance are costly, leading to insufficient coverage in extensive areas \cite{li2014spatial}. 
To compensate for these coverage gaps,  a common practice is to deploy numerous low-cost miniature sensors \cite{chen2017open, chen2022exposure, enigella2018real,mahajan2021yourself}. 
The cost of one AQMS is above one hundred times the cost of one microsensor. Thus, using the same budget, the deployment of microsensors might reach a density of observations 100 times higher than the deployment of regulatory AQMSs. 
Figure \ref{fig:intro} from \cite{chen2022exposure} shows distributions of three different types of data sources in Taiwan, including AQMSs, AirBox sensors maintained by Academia Sinica \cite{chen2017open}, and the Taiwan EPA's Air Quality MicroStations \cite{act2011environmental}. 
The quality of the low-cost sensors is not as good as AQMSs. Additionally, the range of attributes collected varies between these sources, with certain pollutants like nonmethane hydrocarbons being exclusively monitored by AQMSs. To fully exploit each source’s unique properties, a specialized approach to multi-source spatial data prediction is needed.

Developing a machine learning model that effectively caters to the intricacies of multi-source spatial point data prediction remains difficult, due to several open challenges: \textbf{1) Difficulty in effectively aligning information across data sources without ground truth.} Data from different sources may exhibit similar trends, yet they often have conflicting information. Each source, with its unique characteristics, could potentially complement another. In the absence of ground truth data,  developing a robust and effective training strategy that can align these varied sources is crucial for maximizing their collective utility.  
\textbf{2) Difficulty in aligning information across various data source qualities.}  The sensor network, comprised of sensors with varying qualities, poses a challenge in aligning information. It is essential to quantify the quality of different sources accurately to prevent the dilution of high-quality data with noise from lower-quality sources. Despite having some prior knowledge about the accuracy of certain sources, quantifying these quality levels and effectively utilizing the data from lower-quality sources remain challenging.  
\textbf{3) Difficulty in aligning information across different spatial locations.} 
The locations of samples are usually different across data sources. Real-world spatial points usually exhibit spatial autocorrelation. However, existing methods struggle to capture the complex spatial relationship patterns when samples come from different locations and data sources. An ideal method should account for the relative positions of all neighbors, adapting to the dynamic influences of the surrounding environment.

To address these challenges, we introduce a comprehensive framework for multi-source spatial point data prediction. This framework is designed with specific components to tackle each identified challenge: To address the first challenge, we propose a self-supervised strategy derived from maximizing the mutual information between the model estimation and the target variable across each data source. This approach aims to align data from all sources in a self-supervised manner, leveraging their collective strengths for enhanced prediction accuracy. 
To address the second challenge, we introduce the concept of a fidelity score which is derived from the likelihood that the target variable in a data source aligns with the ground truth. This parameter is unconstrained and designed to be learnable, allowing the model to adjust and quantify the quality of each data source dynamically using gradient descent algorithms. 
For the third challenge, we go beyond existing methodologies by developing a geo-location-aware multi-source graph neural network. This network is uniquely equipped with shared and distinct modules to process heterogeneous types of features, thereby effectively capturing the complex spatial relationships across different data sources. Our contributions are thus encapsulated as follows:
\begin{itemize}[leftmargin=*]
  \item We propose the Deep Multi-source Spatial Prediction (DMSP) framework that aligns diverse data sources in a self-supervised manner.
  \item {We introduce the concept of fidelity score, a learnable parameter for quantifying the quality of each data source, enabling the effective utilization of mixed-quality data.} 
  \item We propose a geo-location-aware multi-source graph neural network designed to handle the complexities of spatial relationships and feature heterogeneity across different data sources. 
  \item We conduct extensive experiments on two public real-world datasets and one synthetic dataset to validate the superiority of our method. 
\end{itemize}

\section{Related Work}
In recent years, various methods have emerged for multi-source spatial point prediction, broadly categorized into three groups:
1) Traditional models. The traditional theory-based models are often based on in-depth domain knowledge, such as those used in air pollutant modeling \cite{liu2021data, petroni2020hazardous,turnock2020historical}. They offer interpretability but face limitations due to computational demands and the quality of available data sources \cite{sun2022graph}. In complex scenarios like river network modeling \cite{schmidt2020microplastic}, the multitude of influencing factors makes mechanistic characterization challenging. A critical drawback of these models is their lack of generalizability across different domains. Some works \cite{xie2017review, saini2020comprehensive} employ meteorological principles and mathematical methods to simulate the chemical and physical processes of pollutants, facilitating prediction. While these models are underpinned by robust theoretical foundations, their specificity to certain domains poses a significant limitation in adaptability to diverse scenarios.
2) Gaussian process (GP) based models, which estimate unknown data points through weighted averages of known values \cite{dumitru2013comparative}. GP applies the idea of random variables to the unknown as well as the values of the neighbor samples \cite{oliver1990kriging}. The unbiasedness and minimum variance constraints ultimately result in solving an optimization problem by adding constraints \cite{ryu2002kriging}. GP methods have shown success in scenarios with strong correlations among data sources \cite{deringer2021gaussian, liu2020gaussian}. These methods require manual selection of distance metrics and kernel functions, parameterized by hyperparameters, to construct a covariance matrix.  The limitation of GP lies in its intrinsic hypothesis in terms of the theoretical variogram and constant in mean assumption. The computational intensity and generation of the unsmooth surface also make it hardly generalizable \cite{wojciech2018kriging}. The simplicity of the kernel equation often falls short in capturing complex nonlinear interactions. Numerical computation errors can lead to ill-conditioned matrices, hindering the accurate solution. Recent advancements, such as NARGP \cite{perdikaris2017nonlinear}, combine machine learning with GP for enhanced flexibility and fit, representing the current state-of-the-art in GP methods for the multi-source prediction problems \cite{cheng2023machine, chen2021recent, cvetek2021survey}.
3) Machine learning models. Compared to traditional and GP methods, machine learning models have gained popularity as a more flexible and computationally efficient approach for learning regression formulas for auxiliary attributes. With increasingly sophisticated architectures, machine learning models have become adept at capturing complex data correlations. For instance, Random Forests have been applied to multi-source spatial problems, utilizing geographical and ancillary features to build decision trees for prediction \cite{hu2017estimating, zamani2019pm2, geng2020random}. Nonetheless, these models often fall short in explicitly learning relationships between spatial points from different sources, limiting their effectiveness in multi-source problems. The trend of combining multiple models to leverage their respective strengths and mitigate limitations is gaining traction \cite{liu2021applicationnn, liu2021application}. In spatial relationship studies, methods like Space2Vec \cite{mai2020multi} encode absolute positions and spatial relationships, emerging as state-of-the-art in spatial prediction \cite{mai2023opportunities, liu2022review}.

\section{Problem Formulation}\label{sec:problem form}
We formalize a point in a 2D geographic space as $s\in \mathbb{R}^2$ and consider $N$ distinct data sets $\{D^{(i)}\}_{i=1}^N$, encompassing varied sources like ground stations and mini sensors. Each set  $D^{(i)}$ is characterized by a unique set of geo-locations $S^{(i)}=\{s^{(i)}_{j}\}_{j=1}^{n_i}$, where $n_i$ is the number of locations in $D^{(i)}$. 
These datasets collectively represent multi-source data within the same geographical region, providing diverse perspectives and measurements of the environmental conditions.
Each geo-location $s^{(i)}_{j}$ within dataset $D^{(i)}$ is linked with a point sample $ ({x}^{(i)}_{j},y^{(i)}_{j})$, where ${x}^{(i)}_{j} \in \mathbb{R}^{p_i}$ represents the ancillary attributes, and $p_i$ is the count of these attributes for $D^{(i)}$. The target observations $\{y^{(i)}_{j}\}_{j=1}^{n_i}$ in $D^{(i)}$ are seen as observed samples from the corresponding random variable $Y^{(i)}$, which is an approximation of the ground truth target variable $\tilde{Y}$. Thus, each dataset is represented as $D^{(i)}=\{({x}^{(i)}_{j},y^{(i)}_{j}|s^{(i)}_{j})\}_{j=1}^{n_i}$.  We denote $y^{(i)} \sim Y^{(i)}$ as a general sampling from $Y^{(i)}$ that may not be included in dataset $D_i$. 
With the preliminary notions, we formalize the multi-source spatial point data prediction problem as: 
\textit{Given multiple sets of data $\{D^{(i)}\}_{i=1}^N$, predict the target variable $\tilde{y} \sim \tilde{Y}$ across the 2D geospace.} 
In our model, for any given random location within the geospatial domain, the prediction generated is denoted as $\hat{y}$, correlating with a random variable $\hat{Y}$ such that $\hat{y} \sim \hat{Y}$. We omit the index for the prediction, reflecting its general applicability across the geospace.
Fundamentally, the core objective of our model can be interpreted as maximizing the mutual information between random variables  $\hat{Y}$ and $\tilde{Y}$, as it would enable our model to make highly accurate predictions by precisely mapping the predicted values to the actual, albeit unobserved, ground truth values.

This task faces several key challenges: 1) Aligning without supervision. The absence of ground truth makes it challenging to supervise and align the predictive mapping from various data sources to the prediction target. It is impossible to simply average across different sources due to their different measurement locations and quality, while it is also difficult to learn how to aggregate them because of the lack of ground truth.
2) Aligning across different qualities. The qualities of different sources are usually different yet difficult to quantify. Different data sources usually have different distributions of measurement errors, which are crucial to be considered while aggregating multi-source data. 3) Aligning across different locations. Different sources have different measurement locations so fusing such data requires not only aggregating across sources but also geolocations. 


\section{Methodology}
\begin{figure*}[htb]
  \centering
  \includegraphics[width=\textwidth]{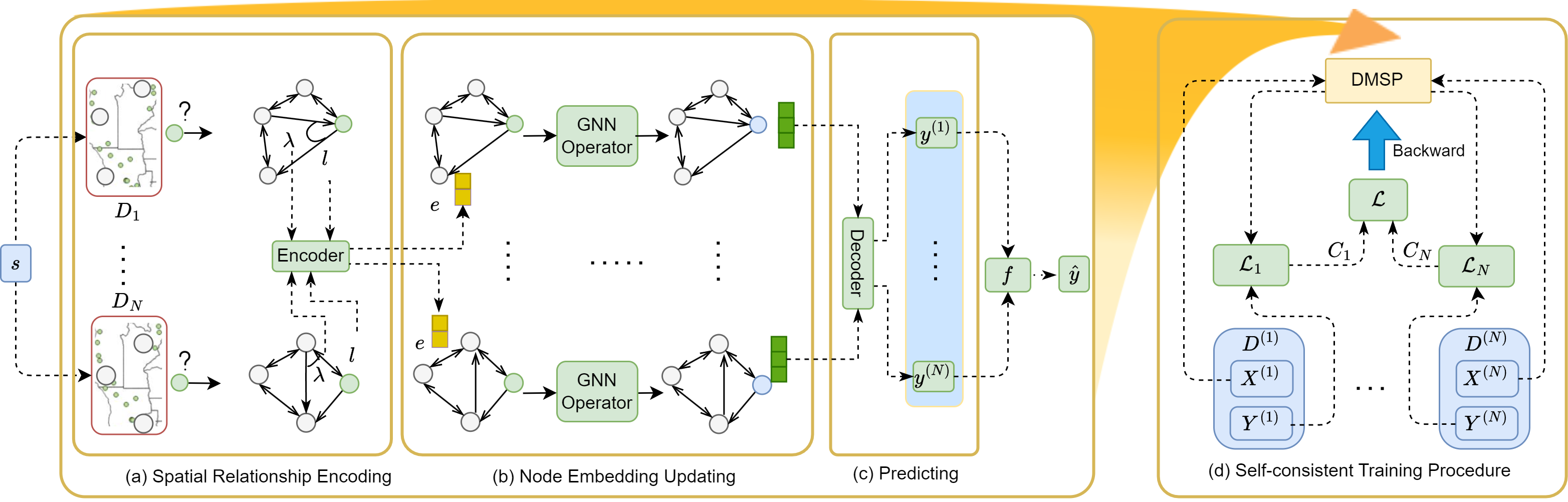}
  \caption{Illustration of the DMSP framework. For a target location $s$, the process involves: (a) building a KNN graph for each data source and using a shared spatial relationship encoder for spatial representations as edge features;  (b) applying distinct graph convolution operators for updating node representations per data source; (c) employing a shared decoder for outputting predictions, which are fused by $f$ for the final prediction. (d) shows the self-consistent training procedure of the framework. }
  \label{fig:arch}
\end{figure*}
To make accurate predictions by addressing the challenges outlined above, we propose the Deep Multi-source Spatial Prediction (DMSP) framework, as shown in Figure \ref{fig:arch}. This framework is designed to align information from diverse data sources. The alignment process is grounded in a self-supervised manner, which is derived from maximizing mutual information. The derivation and details of this approach are thoroughly discussed in Section \ref{sec:mutual}. A key component in assessing the quality of each data source within this framework is the concept of a fidelity score. This score quantitatively represents the quality of each data source. We delve into the specifics of the fidelity score, including its definition and role in the training procedure, in Section \ref{sec:fidelity}. This section also offers a brief overview of the DMSP framework. In Section \ref{sec:GNN},  we focus on the intricate details of the DMSP framework. Specifically, we explore the individual components that constitute DMSP, including independent graph convolution operators, a shared spatial relationship encoder, and a shared decoder.

\subsection{Adaptive Self-supervised Fusion of  Multiple Data Sources}\label{sec:mutual}
As mentioned in the problem formulation section our objective is maximizing the mutual information between $\hat{Y}$ and $\tilde{Y}$.
However, a crucial challenge arises from the unavailability of the ground truth $\tilde{Y}$. To address this, we propose a new idea. Specifically, we only leverage those ground truth values that are well observed by any sensors and by definition of weighted mutual information:
\begin{equation}\label{eq:mutual0}
    \ MI(\tilde{Y}, \hat{Y})=\sum_{\hat{y} \in \hat{Y}}\sum_{\tilde{y} \in \tilde{Y}} w(\tilde{y},\hat{y}) p(\tilde{y},\hat{y}) \log \frac{p(\tilde{y},\hat{y})}{p(\tilde{y}) p(\hat{y})},
\end{equation}
where $w(\tilde{y},\hat{y})$ is the weight for generalized mutual information. We quantify $w(\tilde{y},\hat{y})$ by the probability of being well estimated by each $y^{(i)}$, and in the extreme case, we only consider the case that the observation equals the ground truth:
\begin{equation}
\begin{split}
    & MI(\tilde{Y}, \hat{Y})=\sum_{i=1}^N \sum_{\hat{y} \in \hat{Y}}\sum_{\tilde{y} \in \tilde{Y}} \mathbbm{1}(y^{(i)}=\tilde{y}) p(\tilde{y},\hat{y}) \log \frac{p(\tilde{y},\hat{y})}{p(\tilde{y}) p(\hat{y})}\\
    &=\sum_{i=1}^N \sum_{\hat{y} \in \hat{Y}} \sum_{{y}^{(i)} \in {Y}^{(i)}} \mathbbm{1}(y^{(i)}=\tilde{y}) p(y^{(i)},\hat{y}) \log \frac{p(y^{(i)},\hat{y})}{p(y^{(i)}) p(\hat{y})},
\end{split}
\end{equation}
where $\mathbbm{1}$ is an indicator function. To generalize to the continuous case, we may use a soft probability to replace the indicator function and have:
\begin{equation}
    \sum_{i=1}^N \sum_{\hat{y} \in \hat{Y}} \sum_{{y}^{(i)} \in {Y}^{(i)}} p(y^{(i)}=\tilde{y}) p(y^{(i)},\hat{y}) \log \frac{p(y^{(i)},\hat{y})}{p(y^{(i)}) p(\hat{y})},
\end{equation}
Here, the probability $p(y^{(i)}=\tilde{y})$ quantifies the overlap between random variables $Y^{(i)}$ and $\tilde{Y}$. Because each data source has its consistent measurement error, we can consider this probability as a constant  $C_i$.  
\begin{equation} \label{eq:cmutual}
    \sum_{i=1}^N \sum_{\hat{y} \in \hat{Y}} \sum_{{y}^{(i)} \in {Y}^{(i)}} C_i p(y^{(i)},\hat{y}) \log \frac{p(y^{(i)},\hat{y})}{p(y^{(i)}) p(\hat{y})}=\sum_{i=1}^N C_i MI({Y}^{(i)},\hat{Y}),
\end{equation}
This equation represents our innovative approach to aligning data from multiple sources in an unsupervised setting. The observations from different data sources serve as approximations, enabling us to harness their correlation with the ground truth target variable.
\begin{equation}\label{eq:pp}
\begin{split}
    &\max \sum_{i=1}^N C_i MI({Y}^{(i)},\hat{Y})= \sum\nolimits_{i=1}^{N} C_i\left( H(Y^{(i)})-H(Y^{(i)} \vert \hat{Y} ) \right) \\ 
    \Leftrightarrow &\max -\sum_{i=1}^{N} C_i H(Y^{(i)} \vert \hat{Y}) =  \sum\nolimits_{i=1}^{N} C_i \mathbb{E}_{p(y^{(i)},\hat{y})}\log{p(y^{(i)} \vert \hat{y})}, 
\end{split}
\end{equation}
In Equation \ref{eq:pp}, the entropy term $H(Y^{(i)})$ is a constant for each dataset. Thus, maximizing mutual information essentially translates to maximizing the weighted sum of the expectation of the log conditional likelihood, which measures the uncertainty of the target random variable $Y^{(i)}$ given the model estimation. An ideal model should precisely determine the target variable. 
The actual conditional distribution $p$ in Equation \ref{eq:pp} is unknown. However, we may let the prediction model learn a distribution $q$ that approximates the true conditional distribution $p$.
According to the non-negativity of KL divergence, we can demonstrate that the conditional entropy $H(Y^{(i)} \vert \hat{Y})$ is minimized when the estimated conditional distribution aligns with the actual distribution. By approximating $p$ with the learned distribution $q$, we establish a lower bound on the expectation of the log conditional likelihood as $\mathbb{E}_{p}\log{q(y^{(i)} \vert \hat{y})}$
\begin{thm}\label{thm1} For two random variables $Y^{(i)}, \hat{Y}\in \mathbb{R}$, any variational approximation of the conditional distribution $p_{Y^{(i)} \vert \hat{Y}}$ will increase the conditional entropy $H(Y^{(i)}\vert \hat{Y})$.
\end{thm}
This theorem supports our strategy of maximizing the lower bound as a means to achieve the objective in Equation \ref{eq:pp}. The optimization problem, as initially formulated in Equation \ref{eq:mutual0}, then becomes:
\begin{equation}\label{eq:finalmax}
    \max \sum\nolimits_{i=1}^{N} C_i \mathbb{E}_{p(y^{(i)},\hat{y})} \log{q(y^{(i)} \vert \hat{y})}.
\end{equation}
In this framework, the discrepancy between the estimation $\hat{Y}$ and the target variable $Y^{(i)}$ is characterized by a chosen distribution, such as Gaussian noise.  
Assuming the learned distribution $q$ follows a normal distribution: $q \sim\mathcal{N}(\hat{Y},\mathbb{E}_{p}(Y^{(i)}- \hat{Y})^2)$. 
Then Equation \ref{eq:finalmax} is equivalent to minimizing a weighted square error loss.  In a general case, we have:
\begin{equation} \label{eq:minloss}
\min \sum\nolimits_{i=1}^{N} C_i \mathcal{L}_i (Y^{(i)}, \hat{Y}),
\end{equation}
where $\mathcal{L}_i (Y^{(i)}, \hat{Y})$ is the tailored loss function for the two random variables. This formulation allows for flexibility in assuming any appropriate noise distribution as per the specific requirements of the problem. The detailed training process to achieve the objective outlined in Equation \ref{eq:minloss} is elaborated in Section \ref{sec:fidelity}.

\subsection{Training Procedure and Framework }\label{sec:fidelity}
\subsubsection{Self-consistent Training Procedure}
As formulated in Equation \ref{eq:minloss}, we transform the mutual information objective into a calculable function suitable for self-supervised training.   Specifically, although direct access to ground truth values is not available, we may use target variable observations from all data sources as proxies for training the model. 
Notably, each loss term in Equation \ref{eq:minloss} is weighted by a constant $C_i$. While we have some prior knowledge about the relative quality of different data sources, the exact values of these constants are unknown.  We define these as fidelity scores for each data source and treat them as learnable parameters. The self-consistent training procedure is designed to align model predictions with observations from all data sources by iteratively masking the target variable.
To be specific, for a given sample $({x}^{(i)}_{j},y^{(i)}_{j}|s^{(i)}_{j})$ from dataset $D^{(i)}$, we mask the target variable $y^{(i)}_{j}$ and obtain a partially masked $D'^{(i)}$. Then the DMSP framework $\Phi_{W}$ estimates $y^{(i)}_{j}$ based on the collection of datasets $\{D^{(1)},\dots,D'^{(i)},\dots ,D^{(N)}\}$, where $W$ represents all parameters of DMSP. The loss $\mathcal{L}$ is calculated based on the prediction and $y_{j}^{(i)}$ according to the specific distribution assumption.  
To ensure meaningful fidelity scores and prevent them from becoming zero or negative, we impose two constraints: firstly, the fidelity score must be non-negative; secondly, the sum of all fidelity scores should total a constant, chosen to be 1 for simplicity. The modified objective in Equation \ref{eq:minloss} is thus:
\begin{equation}\label{eq:3st}
         \min_{ {W; \{{C}_i\}_{i=1}^{N} }}  \sum\limits_{i=1}^{N}\sum\limits_{j=1}^{n_i} C_i \mathcal{L}_i(y^{(i)}_{j}, \Phi_{W})
        \; \text{s.t.} \; 0\leq C_i, \sum\limits_{i=1}^{N} C_i=1.    
\end{equation}     
Given the two constraints, we cannot optimize the parameters using standard gradient descent. However, we address this by redefining the fidelity score $C_i$ as a function of an unconstrained variable  $\overline{C}_i$, where $C_i=\frac{\exp{\overline{C}_i}}{\sum_{k=1}^N \exp{\overline{C}_k}}$.

\begin{thm}\label{thm2}Optimizing the unconstrained variables $\{\overline{C}_i\}_{i=1}^N$ is equivalent to optimizing the constrained variables $\{{C}_i\}_{i=1}^N$.
\end{thm}
This theorem allows us to transform our constrained optimization problem into an unconstrained one:
\begin{equation}\label{eq:14}
\min_{ {W; \{\overline{C}_{i=1}^N\} }} \sum\nolimits_{i=1}^{N}\sum\nolimits_{j=1}^{n_i}  \frac{\exp{\overline{C}_i}}{\sum_{k=1}^N \exp{\overline{C}_k}} \mathcal{L}(y^{(i)}_{j}, \Phi_{W} ). 
\end{equation}
The details of the self-consistent training procedure are illustrated in Algorithm \ref{algo}. Line 1 iterates through all datasets.  Line 2 iterates through all samples of the current dataset. Line 3 gets the $j$-th sample. Line 4 masks the target value in the sample to obtain a temporal dataset. Line 5 predicts the masked target value using the proposed DMSP framework. Line 6 calculates the training loss. Line 7 to Line 10 updates the fidelity scores and the parameters of DMSP. 
\begin{algorithm}
\SetAlgoLined
\KwIn{ $\{D^{(i)}\}_{i=1}^N$, initialized $\{C_i\}_{i=1}^N$,  initialized $\Phi_{W}$,  Learning rate $\alpha$.}
\KwOut{$\{C_i\}_{i=1}^N$, $\Phi_{W}$.}

\For{$i \in \{1,\dots, N\}$ }{

\For{ $j \in \{1,\dots,n_i\}$}{
     Get $({x}^{(i)}_{j},y^{(i)}_{j}|s^{(i)}_{j})$\;
  $D'_i \gets D^{(i)}$ with $y^{(i)}_{j}$ masked\;
$\hat{y} \gets \Phi_{W}(D^{(1)},\dots,D'^{(i)},\dots ,D^{(N)})$ \;
$\mathcal{L} \gets C_i \mathcal{L}_i(\hat{y}, y^{(i)}_{j})$ \;
    \For{ $C \in \{C_1,\dots, C_N\}$}{
        $C \gets C -\alpha  \frac{\partial\mathcal{L}}{\partial C}$\;
    }
    ${W} \gets W-\alpha \frac{\partial\mathcal{L}}{\partial W}$\;
  }
 }
 \Return{$\{C_i\}_{i=1}^N$, $\Phi_{W}$}
 \caption{Single epoch training procedure}\label{algo}
\end{algorithm}

\subsubsection{Deep Multi-source Spatial Prediction} \label{sec:prediction model} 

To make predictions by fusing information from all data sources, we propose the DMSP framework as shown in Figure \ref{fig:arch}. For a given target location $s$, we first construct a K-Nearest Neighbor (KNN) graph for each data source based on sample geo-locations (detailed in Section \ref{sec:graph build}).  To consider the influences of geolocation environments, we propose a spatial relationship encoder $\zeta_{SD}$ (Figure \ref{fig:arch} (a)) to extract the spatial relationships from the coordinate values. The encoder is shared across all data sources due to their shared geographic space (Figure \ref{fig:arch} (a)). Details of the encoder are elaborated in Section \ref{sec:rep}.  

After extracting spatial representations, distinct graph convolution operators are applied to each data source to fit their specific attribute and target variable relationships, as shown in Figure \ref{fig:arch} (b). These operators facilitate specialized training for unified node embeddings, essential for target variable prediction (detailed in Section \ref{sec:GNN MP}).

In Figure \ref{fig:arch} (c), a shared decoder $\zeta_{SD}$, which can be a multilayer perceptron network, transforms the final representation of the target node into the target variable prediction for each data source.  Finally, the predictions from different sources are fused through a prediction function $f$. In practice, we choose to implement $f$ as a weighted sum of all source predictions based on their fidelity scores: $\hat{y}= \sum_{i=1}^N C_i {y}^{(i)}$. In Figure \ref{fig:arch} (d), fidelity scores are utilized to calculate weighted losses during training with target variables from different sources. By utilizing a gradient descent algorithm, our framework optimizes towards the objective in Equation \ref{eq:minloss}.

\subsection{Geo-location-aware Graph Neural Networks} \label{sec:GNN}
In this section, we introduce the construction of the input graph for prediction and the integration of spatial information into the graph neural network within the DMSP framework.  
\subsubsection{Spatial K-Nearest Neighbor Graph} \label{sec:graph build}
Spatial data inherently exhibits spatial correlation,  where nearby points often share similar attributes.  To harness this characteristic, we construct a K-nearest neighbor (KNN) graph based on sample locations. For a sample $({x}^{(i)}_{j},y^{(i)}_{j}|s^{(i)}_{j})$ in data source $D^{(i)}$, with $s^{(i)}_{j}$ representing the locations of simultaneous samples, we build a KNN graph $G_{j}^{(i)}$ using $s^{(i)}_{j}$ and $s^{(i)}_{j}$. 
It's important to note that samples within the same data source may have different timestamps, leading to a varying number of samples ($n_{i,j}$) at a given time. Thus $n_{i,j}$ is less than the total sample number $n_i$ in data source $D^{(i)}$. 

The graph $G_{j}^{(i)}= (V_{j}^{(i)}, E_{j}^{(i)}, \overline{x}^{(i)}_{j})$ consists of a node set $V_{j}^{(i)}$, which is the union of the target location $s^{(i)}_{j}$ and $s^{(i)}_{j}$, and edge feature set $E_{j}^{(i)} \subseteq V_{j}^{(i)} \times V_{j}^{(i)}$ denoting the relationships between locations. The edge feature will be detailed in Section \ref{sec:rep}.  Directed edges connect the K-nearest neighbors to the central node. Therefore, there are exactly $K\Vert V_{j}^{(i)}\Vert$ edges in the graph.
The node attribute matrix $\overline{x}^{(i)}_{j}\in \mathbb{R}^{n_{i,j} \times (p_i+1)}$ corresponds to features from the associated samples.

\subsubsection{Spatial Relationship Encoder}\label{sec:rep}
Capturing spatial relationships between nodes in a graph requires more than just using coordinate values or distances, as these methods often fall short in integrating comprehensive spatial information. Coordinates can impose unnecessary constraints and fail to recognize simple translations or rotations within the dataset. Similarly, distances alone cannot represent the orientation of neighboring nodes or account for the combined effects of multiple neighboring nodes. Recognizing these limitations, we adopt a more holistic approach.

Instead of relying on a simplistic distance-based adjacency matrix, our approach jointly considers distances and angles to represent node relationships.  We utilize a spatial relationship encoder $\zeta_{SE}$  to dynamically compute the relationships between nodes based on both distance and angular measurements.  
For each edge $(v_{j;k}^{(i)}, v_{j;o}^{(i)})$ in the graph, where $v_{j;o}^{(i)}$ is the target node and $v_{j;k}^{(i)}$ is a source node, we define an edge representation $e_{j;k,o}^{(i)} \in E_{j}^{(i)}$ as follows:
\begin{equation}\label{eq:rep_k}
    e_{j;k,o}^{(i)}=\zeta_{SE}\left (l_{j;o,k}^{(i)} ,  \lambda_{j;o,k}^{(i)}\right),
\end{equation}
where $l_{j;o,k}^{(i)}$ is the distance between the two nodes, and  $\lambda_{j;o,k}^{(i)}\in [-\pi, \pi)$ is the angle at node $v_{j;k}^{(i)}$ formed by rotating clockwise from the ray $\overrightarrow{v_{j;k}^{(i)} v_{j;o}^{(i)}}$ until it aligns with the direction of the next neighboring node of $v_{j;k}^{(i)}$.
This representation is generalized from  \cite{yu2022deep}. Compared to the original representation, we have a relaxed requirement for the number of samples as we do multi-source prediction. 
\subsubsection{Graph Convolution Operators}\label{sec:GNN MP}
To derive meaningful node embeddings that capture attribute-target relationships by incorporating the computed spatial representations, we employ multiple layers of graph convolutional operators. These operators process both node and edge features. The updated node attributes are expressed as
$\overline{x}^{(i)(l+1)}_{j}=g^{(i)(l)}(\overline{x}^{(i)(l)}_{j},E_{j}^{(i)})$,
where $\overline{x}^{(i)(l+1)}_{j}$ represents the updated node attributes after $l$ layers, and $g^{(i)(l)}$ is the graph convolution operator at layer $l$ for graphs constructed from data source $D^{(i)}$. It's important to note that these graph convolutional operators are distinct for each data source, allowing for customized processing of different data characteristics.
After applying $L$ layers of graph convolution, the embedding of the target node is ${x}_{j;k}^{(i)(L)}$, which is passed to the downstream decoder to make the final prediction.
The dynamic nature of the graph $G_{j}^{(i)}$, which changes over time and adapts to varying sensor data, enables the model to learn inductively, without reliance on a fixed graph structure. 
\section{Experiment}
In this section, we first introduce the experimental settings, then we evaluate the effectiveness and robustness of the proposed method and show the performance result against comparison methods. After that, we perform ablation studies to verify the contribution of different parts of the proposed framework. Finally, we show some analysis of scalability and the sensitivity of hyperparameters to help further explore the characteristics of the proposed framework. The implementation details and efficiency analysis can be found in the appendix. All the experiments are conducted on a 64-bit machine with an NVIDIA A5000 GPU.
\subsection{Experiment Setting}
\subsubsection{Dataset}
1) \textbf{SouthCalAir}\cite{singh2021sensors}: Fine particulate matter (PM 2.5) data from 26 Environmental Protection Agency's (EPA) Air Quality System (AQS) sensors and 515 PurpleAir sensors in south California \cite{snyder2013changing} from 2019/01/01 to 2019/12/31. EPA AQS data are considered high-quality while PurpleAir data are noisy.
2) \textbf{NorthCalAir}\cite{singh2021sensors}: Similar to the SouthCalAir dataset, PM 2.5 data from 63 Environmental Protection Agency's (EPA) Air Quality System (AQS) sensors and 1110 PurpleAir sensors in north California from 2019/01/01 to 2019/12/31.
3) \textbf{Spatially correlated regression (SCR)}: A synthetic dataset generated by simulating spatially correlated data that involves creating features distributed across space with spatial correlations \cite{getis2009spatial} and simulating non-linear relationships among features and the target variable \cite{leung2000statistical}. Gaussian normal noise is added to the target variable to simulate the low-quality data source. Features are irreversibly transformed to avoid degradation to a simple regression task.
4) \textbf{Flu}: The Flu dataset integrates data from two distinct sources: the Centers for Disease Control and Prevention (CDC) and the Google Flu Trends program \cite{dugas2013influenza} from 2010 to 2015. The CDC provides data from medical providers, which is considered more reliable but is limited in geographical coverage (48 states). Conversely, Google Flu Trends estimates activity levels based on public keyword searches, offering broader coverage but less reliability. The results of this dataset can be found in the appendix.




\subsubsection{Comparison Method}
1) \textbf{SRA-MLP} \cite{liu2021applicationnn}: A model that combines the stepwise regression analysis and multilayer perceptron neural network. 
2) \textbf{RR-XGBoost} \cite{liu2021application}: A model that combines ridge regression and the Extreme Gradient Boosting Algorithm.
3) \textbf{NARGP} \cite{perdikaris2017nonlinear}: A probabilistic framework based on Gaussian process regression and non-linear autoregressive schemes.
4) \textbf{GeoPrior} \cite{mac2019presence}:  An efficient spatial prior is proposed in the method to estimate the probability conditioned on a geographical location.
5) \textbf{Space2Vec} \cite{mai2020multi}: A representation learning model to encode the absolute positions and spatial relationships of places.   
\begin{table*}
\renewcommand{\arraystretch}{1}
\small
  \caption{The performance of the proposed model (including ablation variants) and the comparison methods.}
  \centering
  \begin{tabular}{c|c|c|c|c|c|c}
    \toprule
    Dataset & Method   & MAE   & RMSE  & EVS & CoD     & Pearson\\
    \midrule
    SouthCalAir
    & SRA-MLP &   \underline{3.211$\pm$0.059} & \underline{5.305$\pm$0.091} & \underline{0.416$\pm$0.017} & \underline{0.411$\pm$0.015} & \underline{0.686$\pm$0.012}\\
    & RR-XGBoost &  5.811$\pm$0.047 & 8.450$\pm$0.107 & -0.258$\pm$0.050 & -0.2644$\pm$0.056 & {0.351$\pm$0.010}\\
        & NARGP  &   4.476$\pm$0.853 & 7.000$\pm$1.484 & 0.084$\pm$0.334 & 0.076$\pm$0.331 & 0.487$\pm$0.138\\
        & DMSP &  \textbf{3.112$\pm$0.059} & \textbf{4.878$\pm$0.234} & \textbf{0.542$\pm$0.026} & \textbf{0.504$\pm$0.036} & \textbf{0.737$\pm$0.021} \\
        \cline{2-7}
        & GeoPrior &   3.236$\pm$0.305 & 5.926$\pm$0.254 & 0.411$\pm$0.107 & 0.411$\pm$0.109 & 0.670$\pm$0.041\\
        & Space2Vec &   3.135$\pm$0.303 & 5.996$\pm$0.283 & 0.401$\pm$0.122 & 0.395$\pm$0.107 & 0.672$\pm$0.056\\
       & DMSP-H &   4.091$\pm$0.486 & 6.106$\pm$0.344 & 0.277$\pm$0.119 & 0.263$\pm$0.139 & 0.555$\pm$0.059\\
        & DMSP-F &   14.835$\pm$2.850 & 22.445$\pm$3.768 & -6.972$\pm$2.271 & -9.361$\pm$2.379 & 0.055$\pm$0.074\\
    \hline
    NorthCalAir
         & SRA-MLP &  \underline{3.000$\pm$0.059} & \underline{5.374$\pm$0.378} & \underline{0.404$\pm$0.026} & \underline{0.390$\pm$0.024} & \underline{0.636$\pm$0.021}\\
        & RR-XGBoost &  3.705$\pm$0.030 & 6.177$\pm$0.334 & {0.121$\pm$0.066} & 0.119$\pm$0.068 & {0.557$\pm$0.014}\\
        & NARGP  &   3.317$\pm$0.035 & 6.26$\pm$0.580 & 0.186$\pm$0.158 & 0.185$\pm$0.158 & 0.579$\pm$0.052\\    
         & DMSP &  \textbf{2.423$\pm$0.083} & \textbf{4.474$\pm$0.489} & \textbf{0.590$\pm$0.052} & \textbf{0.586$\pm$0.046} & \textbf{0.768$\pm$0.034} \\ 
         \cline{2-7}
        & GeoPrior &   2.845$\pm$0.055 & 4.920$\pm$0.154 & 0.481$\pm$0.027 & 0.480$\pm$0.029 & 0.690$\pm$0.021\\
        & Space2Vec &   2.641$\pm$0.103 & 4.509$\pm$0.283 & 0.585$\pm$0.022 & 0.584$\pm$0.027 & 0.762$\pm$0.056\\     
        & DMSP-H &   3.768$\pm$0.823 & 6.048$\pm$1.394 & 0.236$\pm$0.274 & 0.221$\pm$0.286 & 0.506$\pm$0.205\\
        & DMSP-F &   17.110$\pm$3.289 & 42.426$\pm$20.589 & -32.214$\pm$33.699 & -37.378$\pm$35.118 & 0.003$\pm$0.013\\      
    \hline
    SCR
        & SRA-MLP &   {0.782$\pm$0.031} & 0.969$\pm$0.029 & 0.007$\pm$0.008 & -0.016$\pm$0.014 & 0.092$\pm$0.080\\
        & RR-XGBoost & 0.939$\pm$0.014 & 1.208$\pm$0.032 & -0.573$\pm$0.177 & -0.646$\pm$0.236 & 0.031$\pm$0.040\\
        & NARGP  &  \underline{0.616$\pm$0.045} & \underline{0.773$\pm$0.059} & \underline{0.457$\pm$0.024} & \underline{0.446$\pm$0.025} & \underline{0.698$\pm$0.012}\\     
        & DMSP &  \textbf{ 0.478$\pm$0.032} & \textbf{0.574$\pm$0.025} & \textbf{0.606$\pm$0.046} & \textbf{0.605$\pm$0.045 } & \textbf{0.780$\pm$0.027} \\
        \cline{2-7}
         & GeoPrior &   0.505$\pm$0.035 & 0.600$\pm$0.054 & 0.553$\pm$0.037 & 0.553$\pm$0.039 & 0.731$\pm$0.021\\
        & Space2Vec &   0.498$\pm$0.015 & 0.615$\pm$0.063 & 0.584$\pm$0.031 & 0.580$\pm$0.029 & 0.751$\pm$0.046\\     
        & DMSP-H &   0.516$\pm$0.026 & 0.631$\pm$0.030 & 0.507$\pm$0.135 & 0.489$\pm$0.158 & 0.726$\pm$0.071\\
        & DMSP-F &   {0.484$\pm$0.018} & {0.609$\pm$0.014} & {0.596$\pm$0.048} & {0.588$\pm$0.047} & {0.776$\pm$0.026}\\       
    \bottomrule
  \end{tabular}
  \label{table:performance}
\end{table*}

\subsection{Effectiveness Analysis}
\subsubsection{Evaluation Metrics}
To quantify the effectiveness of the models, we use the following five metrics: 1) Mean absolute error (MAE), 2) Root-mean-square error (RMSE), 3) Explained variance score (EVS),
4) Coefficient of determination (CoD), 5) Pearson correlation coefficient (Pearson).
\subsubsection{Performance}

Table \ref{table:performance} summarized the performance comparison among the proposed methods and competing models on the real-world and synthetic datasets. The results are obtained from 3 individual runs for every setting. We report the standard deviation over three runs following the $\pm$ mark. The best result for each dataset is highlighted with boldface font and the second best is underlined. In general, our proposed framework outperformed comparison methods in terms of all the evaluation metrics.
Note that the GeoPrior and Space2Vec can be included in our framework to substitute the Geo-location-aware graph neural network part, so we include them as ablation study results in this table, together with two other ablation models DMSP-H and DMSP-F, which will be introduced later. 
It is apparent from the results that the proposed method excels over other methods and achieves the best performance on all the datasets. 
For instance, DMSP reduced the MAE error by 27.5\% compared to the average prediction of other methods for the air pollutant prediction task on the NorthCalAir dataset, and by 19.2\% compared to the second-best model. What's more, DMSP achieves a Pearson score of 0.768 while the second-best score is only 0.636 on the NorthCalAir dataset.  For the comparison methods, Even though they can achieve similar scores sometimes, their prediction performance may not be consistent on all the datasets. For example, SRA-MLP delivers the second-best scores on the SouthCalAir and NorthCalAir datasets, but it performs badly on the SCR dataset. This is due to the fact that the target value in the SCR dataset can not be determined by the sample features. SRA-MLP lacks the ability to predict based on the information of neighbors by learning spatial relationships. The GP-based SOTA NARGP has a consistent performance on all the datasets, but the overall performance is not comparable to DMSP. This implies that NARGP has the ability to model spatial correlations and predict according to neighbor samples. However, since it relies on fixed kernels, it fails to accurately model the spatial relationships in the dataset. 

\subsection{Ablation Study}
We conduct three ablation studies here. We first investigate the benefit gained from the proposed objective function in Equation \ref{eq:3st}. Specifically, we 1) examine whether a single data source has enough information; and 2) examine the impact of treating each data source equally. Then we investigate the effectiveness of the proposed graph neural network by 3) comparing it with two SOTA spatial embedding methods.

\subsubsection{Information from a Single Source}
For the first question, we propose to use the target variable in a single source as supervision in the loss calculation. The best result on a single data source is reported and we name this variation of our model DMSP-H. The results on all the datasets are shown in Table \ref{table:performance}. Overall, the performance will degrade if the model only learns from a single data source. Particularly, the performance drop is most significant on the NorthCalAir dataset, and the DMSP-H becomes the worst among the comparison methods. This is due to the fact that it is the largest dataset where each source provides a significant contribution to the final prediction. This study shows that although a single source cannot provide enough information to achieve the best performance. 
\subsubsection{Mixed Multiple Sources}
We further conduct another ablation study where we remove the use of fidelity score and treat all the data sources equally. We name this model DMSP-F and the performances are shown in Table \ref{table:performance}. We notice a huge performance drop on the two air pollutants datasets. This implies that the low-quality sources in the air pollutants datasets are extremely noisy and provide misleading guidance for the whole model. Notably, the performance on the SCR dataset almost does not change. This is due to the fact that we only add a zero-mean Gaussian random noise to the low-quality data source in the SCR dataset. So the general trend of the low-quality data source is correct and the neural network model is resistant to such an ideal noise. This also suggests that the noise in the real-world dataset is hard to model. This study shows the huge noise from real-world datasets and indicates that there is not always a benefit from fusing multiple data sources.
\subsubsection{Different Spatial Embeddings in GNN}
We substitute the proposed geo-location-aware graph neural networks by GeoPrior and Space2Vec. As we can observe, GeoPrior does not perform well. This is because the original coordinate values do not hold any meaningful information for the prediction problem. The Space2Vec also fails to provide meaningful information for the spatial point prediction problem.  In the point data prediction problem,  the crucial geo factor is the relative position and angles between the target node and its neighbors. The two comparison methods aim to create a unique embedding for any location in the study area from a global perspective, similar to the positional encoder method used in language models like transformers.  

\begin{figure*}[htb]
  \centering
  \includegraphics[width=\textwidth]{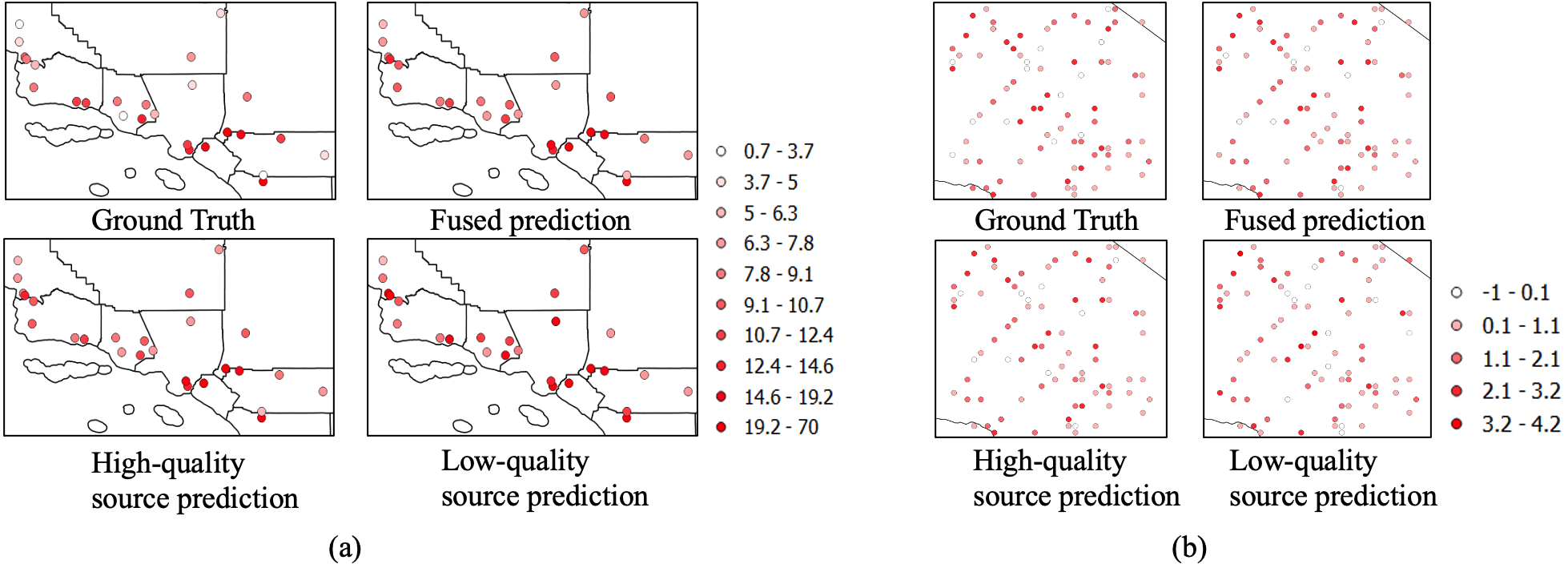}
  \caption{Visualization for the prediction and approximate ground truth for (a) SouthCal and (b) SCR}
  \label{fig:visualization}
\end{figure*}
\subsection{Case Study}
Here we further provide two case studies on the SouthCal and SCR datasets showing the predictions from each data source and the fused prediction compared with the observations from the high-fidelity data source as the approximation of the ground truth. The unconstrained fidelity scores are $\{14.837, -17.930\}$ for SouthCal and $\{2.836, -1.836\}$ for SCR. The result matches the prior knowledge that the low-quality data source is assigned a low score. As shown in Figure \ref{fig:visualization}, in general, the prediction from the high-quality source is closer to the observed value, while the low-quality source prediction is not accurate. And the proposed framework can automatically learn the fidelity scores which assign the weight that can effectively safeguard against wrong estimations from the low-quality data source.



\subsection{Scalability Analysis}
\begin{figure}[!h]
     \centering
     \begin{subfigure}[b]{0.495\linewidth}
         \includegraphics[width=\textwidth]{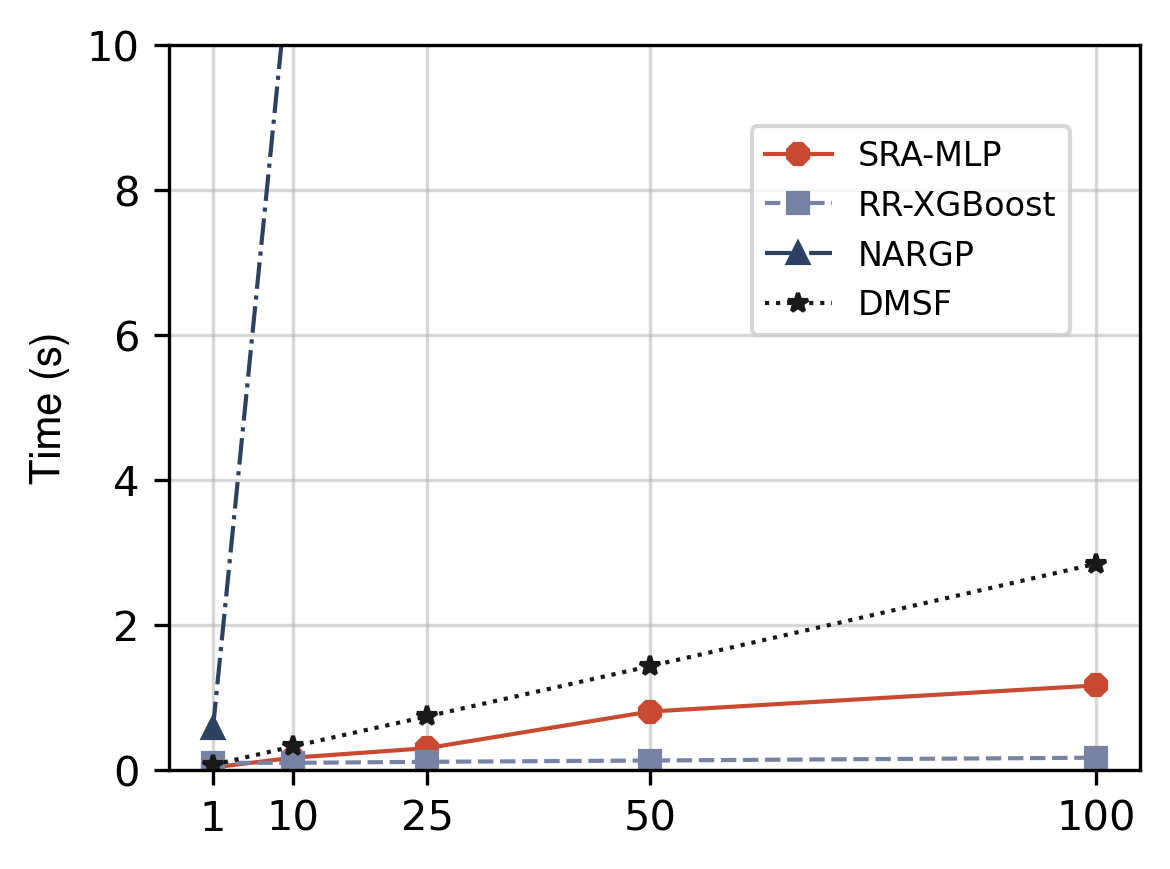}
         \caption{Training}
         \label{fig:scale train}
     \end{subfigure}
     \hfil
     \begin{subfigure}[b]{0.495\linewidth}
         \includegraphics[width=\textwidth]{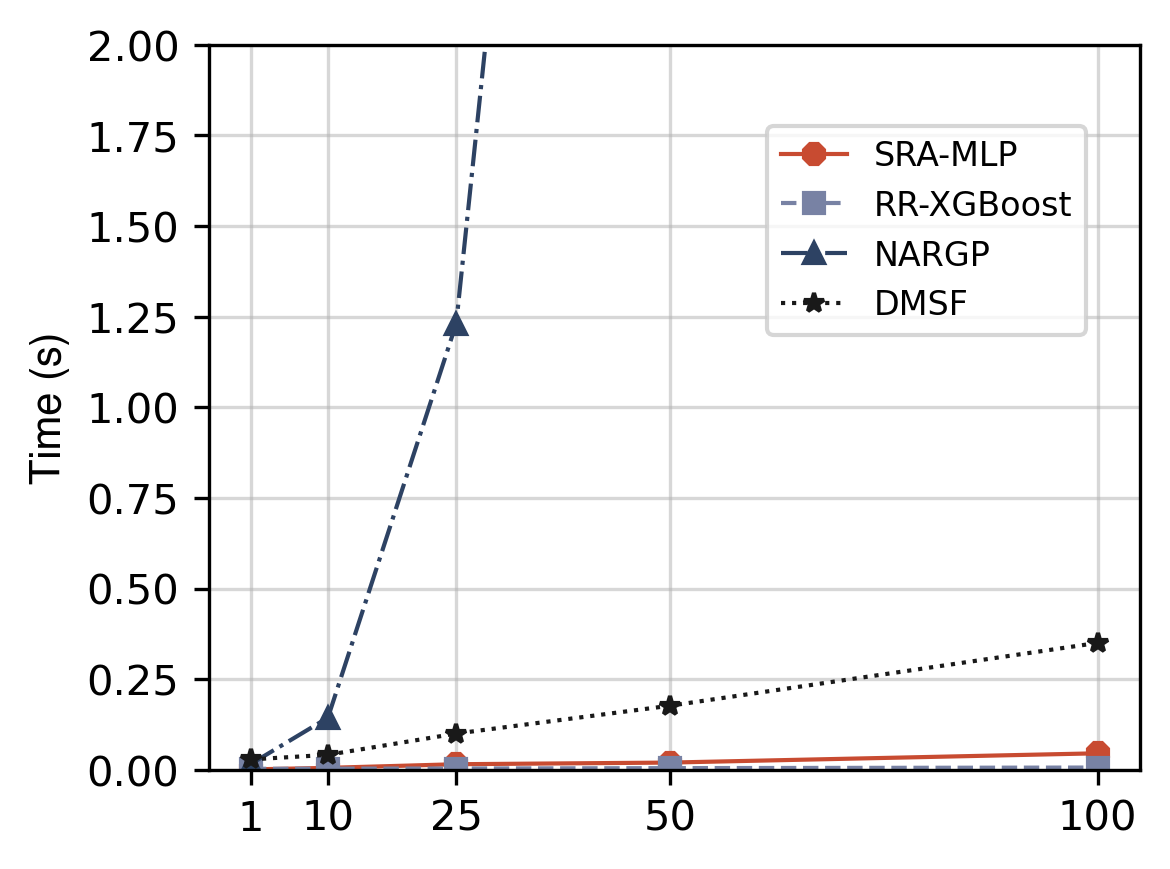}
         \caption{Testing}
         \label{fig:scale test}
     \end{subfigure}
      \caption{The scalability study about the number of samples.}
        \label{fig:scale sample}
\end{figure}

To analyze the scalability of the proposed framework, we record the average runtime for all the methods. The runtime is shown with respect to the number of samples. We take the number of samples in the original SCR dataset as the counting unit and generate 10, 25, 50, and 100 times the number of samples. The results are presented in Figure \ref{fig:scale sample}. As we can see, all the machine learning methods demonstrate similar linear increase patterns with the growth of the number of samples. Our model has a slightly higher runtime than other machine learning methods, but they are still in the same order of magnitude. 
The test case is similar to the training one, where our model still shows linear growth with respect to the number of samples. 

\subsection{Sensitivity Analysis}
\begin{figure}[!htb]
     \centering
     \begin{subfigure}[b]{0.495\linewidth}
         \includegraphics[width=\textwidth]{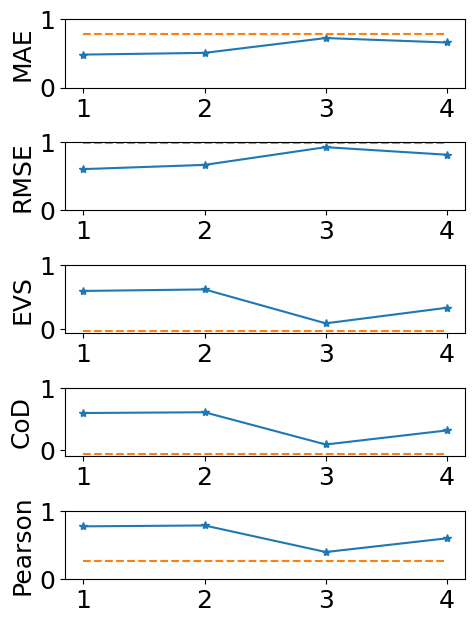}
         \caption{Number of neighbors}
         \label{fig:sensi neighbor}
     \end{subfigure}
     \hfil
     \begin{subfigure}[b]{0.495\linewidth}
         \includegraphics[width=\textwidth]{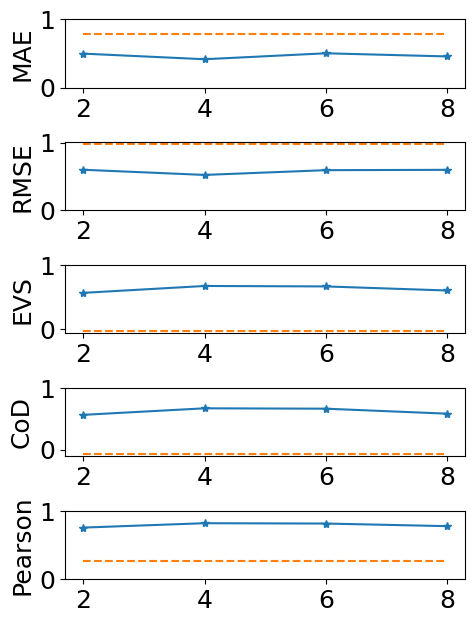}
         \caption{Number of GNN layers}
         \label{fig:sensi layers}
     \end{subfigure}
      \caption{The sensitivity study of two hyperparameters in DMSP on the SCR dataset. The dashed lines represent the average baseline performance.}
        \label{fig:sensitivity}
\end{figure}

Here we provide a sensitivity analysis of two hyperparameters in the proposed framework: the number of neighbors $K$ chosen to build the KNN graph and the number of GNN layers. Figure \ref{fig:sensitivity} shows the performance for various values of the hyperparameters on the SCR dataset.
The sensitivity of the number of neighbors is shown in Figure \ref{fig:sensi neighbor}. In general, the model is not very sensitive to the number of neighbors within the range studied. And it outperformed the baseline by a significant margin in terms of all the metrics.  
The optimal value for the number of neighbors is around 4.
While the specific best number of neighbors can vary depending on the dataset as well as the task, in general, the proposed framework can perform well when the number is relatively small.
The sensitivity of the number of GNN layers is shown in Figure \ref{fig:sensi layers}. 
The optimal value for the number of GNN layers is around 2.

\section{Conclusion}
We propose a new framework for multi-source spatial point data prediction. Central to this framework is a self-supervised training objective that not only aligns data from various sources but also leverages their collective strength without ground truth data. The fidelity score quantitatively evaluates the quality of each data source.  Moreover, the geo-location-aware graph neural network adeptly handles the intricate spatial relationships across different locations. 
The efficacy of our methods has been rigorously tested and validated on both synthetic and real-world multi-source datasets. The results demonstrate that our framework not only outperforms existing state-of-the-art methods but also provides meaningful insights into the data.

\begin{acks}
This work was supported by the NSF Grant No. 2432418, No. 2414115, No. 2007716, No. 2007976, No. 1942594, No. 1907805, No. 2318831, Cisco Faculty Research Award, Amazon Research Award.
\end{acks}

\bibliographystyle{ACM-Reference-Format}
\bibliography{software}

\appendix
\section{Supplementary Material}

\subsection{Proof of Theorem \ref{thm1}}
\begin{proof}
  Let $H'(Y^{(i)} \vert \hat{Y}) $ be the variational approximated conditional entropy,
  \begin{align*}
  H'(Y^{(i)} \vert \hat{Y}) &= -\mathbb{E}_p  \log{q(y^{(i)} \vert \hat{y})} \\
    H(Y^{(i)}\vert \hat{Y}) &= -\mathbb{E}_p \log{p(y^{(i)}\vert \hat{y})} \\
     &= - \mathbb{E}_p   \left[\log{p(y^{(i)}\vert \hat{y})} + \log{q(y^{(i)}\vert \hat{y})} - \log{q(y^{(i)}\vert\hat{y})}\right] \\ 
     &= -\mathbb{E}_p   \left[\log{q(y^{(i)}\vert \hat{y})} +  \log{\frac{p(y^{(i)}\vert \hat{y})}{q(y^{(i)}\vert \hat{y})}}\right] \\
     &= -\mathbb{E}_p \log{q(y^{(i)}\vert \hat{y})} -{D}_{\mathbb{KL}}(p\parallel{q}) \\
    & \leq -\mathbb{E}_p  \log{q(y^{(i)}\vert \hat{y})} \\
     &= H'(Y^{(i)}\vert \hat{Y}) 
  \end{align*}
\end{proof}
\subsection{Proof of Theorem \ref{thm2}}
The proof of this theorem is a consequence of the following two lemmas.
\textbf{Lemma 1.} \textit{$\{{C}_i\}_{i=1}^N$ obtained from $\{\overline{C}_i\}_{i=1}^N$ satisfies the original two constraints on $C_i$.}
\begin{proof}
\begin{align*}
    &0 \leq C_{i} = \frac{\exp{\overline{C}_i}}{\sum_{k=1}^N \exp{\overline{C}_k}} \leq 1, \\
    & \sum_{i=1}^{N} C_i = \frac{\sum_{i=1}^{N} \exp{\overline{C}_i}}{\sum_{k=1}^N \exp{\overline{C}_k}} =1,
\end{align*}
\end{proof}
\textbf{Lemma 2.} \textit{For any optimal set $\{{C}_i^*\}_{i=1}^N$ from the original optimization problem, there exists $\{{C}_i\}_{i=1}^N$ equals $\{{C}_i^*\}_{i=1}^N$.}
\begin{proof}
Base case: For ${C}_1^* \in [0,1]$, we can have $C_1=C_1^*$. 
Inductive hypothesis: Assume we have $\{C_i=C^*_i\}_{i=1}^n, n \geq 1$.
Inductive step: Let $A=\sum_{i=1}^n C_i$. Then $C_{i+1}^* \in [0,1-A]$, and $C_{i+1} \in [0,1-A]$ can take any value in the range, so we can have $C_{i+1}=C_{i+1}^*$. 
By induction, we can have $\forall i \in \{1,\cdots, N\}, C_{i}=C_{i}^*$.  
\end{proof}

\begin{proof} 
As stated in Lemma 1 and 2, if there exists an optimal solution $\{{C}_i^*\}_{i=1}^N$ of the original optimization problem, we can get the same set that satisfies the original constraints via optimizing $\{\overline{C}_i\}_{i=1}^N$. And it's obvious that the set $\{{C}_i\}_{i=1}^N$ obtained from the optimal $\{\overline{C}_i^*\}_{i=1}^N$ of the new problem must also be optimal in the original problem.
\end{proof}

\subsection{Implementation details}
We assume a Gaussian normal noise for the data and choose the RMSE loss for $\mathcal{L}$. For the spatial feature encoder $\zeta_{SE}$ and the decoder $\zeta_{SD}$, we leverage 3-layer MLPs. For each dataset, we randomly split it into 60\%, 20\%, and 20\% for training, validation, and testing respectively. All the Deep learning models are trained for a maximum of 500 epochs using an early stop scheme. Adam optimizer with a learning rate of 0.001 is used. 
For the SRA-MLP method, we leveraged the LinearRegression model in sklearn package and implemented a multi-layer perceptron network using the PyTorch framework. The hyperparameters we tuned and their respective range were: dimension of hidden features in {[32,64,128]}, the number of layers in {[2,4,8,16]}. The best hyperparameter we found were 128 and 8.
For the RR-XGBoost method, we used the Ridge Regression model in the scikit-learn package and the xgboost package to implement the model. We used default values for most of the parameters and focused on two sensitive hyperparameters: number of estimators in {[100,1000,10000]} and max depth in {[2,4,8]}. We found the best hyperparameter were 1000 and 4.
For the NARGP method, since the code is available we followed the original settings but increased the maximum iteration number to 2000 to ensure convergence.
For our DMSP model, we used a 3-layer MLP for the decoder and a single-layer forward network for the spatial relationship encoder. We set the dimension of hidden embeddings in all the modules to be the same (16). The hyperparameters we tuned are the number of neighbors K in {[1,3]} and the number of GNN layers in {[2,4,8]}. We set K to 3, and the number of GNN layers to 2.

\subsection{Performance on Flu dataset}
The results on the Flu dataset are detailed in table \ref{table:flu}. By applying our method, we can synergistically combine these sources to enhance prediction accuracy. Our approach has consistently outperformed other methods by at least 9\% across all metrics, demonstrating its effectiveness in integrating and leveraging multi-source data for superior predictive outcomes.
\begin{table}
\small
  \caption{The performance on flu dataset.}
  \centering
  \begin{tabular}{c|c|c|c|c|c|c}
    \toprule
    Dataset & Method   & MAE   & RMSE  & EVS & CoD     & Pearson\\
    \midrule
    Flu
    & SRA-MLP &   0.062& 0.095 &0.115 & 0.107 &0.351\\
    & RR-XGBoost &  0.744& 0.859& 0.168& 0.168& 0.440\\
        & NARGP & 0.045 &0.066 &0.379 &0.376& 0.620\\
        & DMSP &  \textbf{0.038} & \textbf{0.060} & \textbf{0.536} & \textbf{0.536} & \textbf{0.733} \\ 
    \bottomrule
  \end{tabular}
  \label{table:flu}
\end{table}

\subsection{Efficiency analysis}
Here we compare the average training and testing runtime per epoch. For statistical methods that do not have epoch training, we record the optimization time for one iteration. Some methods need to do data pre-processing before they can make predictions, and this procedure can be saved into the dataset. To make a fair comparison, we only record the time used for forward and backward operations for the same number of predictions.  As shown in Table \ref{table:eff}, the training time of our proposed method is between the two machine learning models and is smaller than NARGP by one order of magnitude. For the testing phase, our model has a slightly higher runtime than the NARGP method.  
\begin{table}[ht]
    \renewcommand{\arraystretch}{0.8}
  \caption{ Running time for Training and Testing}
  \label{table:eff}
  \centering
  \begin{tabular}{c c c c c }
    \toprule
    Methods &  SRA-MLP    & RR-XGBoost   &  NARGP  &   DMSP \\
    \midrule
        Training & 0.0325s & 0.0963s & 0.5732s & 0.0685s \\
        Testing & 0.0011s & 0.0028s & 0.0163s & 0.0201s\\
    \bottomrule
  \end{tabular}
\end{table}


\end{document}